\documentclass{llncs}

\usepackage{wrapfig}
\usepackage{llncsdoc}
\usepackage{epsfig,psfrag}
\usepackage{latexsym,amssymb,amsmath}
\usepackage{longtable}
\usepackage{tikz,pgf}
\usepackage{subfig}
\usepackage{wrapfig}
\usepackage{rotating}

\usepackage{graphicx}
\usepackage{epstopdf}
\usepackage{color}

\usepackage{amsmath,amsfonts}
\newcommand{\bqn}{\begin{eqnarray}}
\newcommand{\eqn}{\end{eqnarray}}
\newcommand{\bq}{\begin{eqnarray*}}
\newcommand{\eq}{\end{eqnarray*}}

\usepackage{url}
\usepackage{hyperref}
\hypersetup{colorlinks,%
citecolor=black,%
filecolor=black,%
linkcolor=red,%
urlcolor=blue,%
pdftex}

\begin{document}

\title{Lattice Paths for Persistent Diagrams}

\author{Moo K. Chung$^1$ \and Hernando Ombao$^2$}
\institute{
$^1$University of Wisconsin, Madison, USA\\
$^2$King Abdullah University of Science and Technology, Thuwal, Saudi Arabia\\
\email{mkchung@wisc.edu, hernando.ombao@kaust.edu.sa}
}

\maketitle

\begin{abstract}
Persistent homology has undergone significant development in recent years. However, one outstanding challenge is to build a coherent statistical inference procedure on persistent diagrams. In this paper, we first present a new  lattice path representation for  persistent diagrams.  We then develop a new exact statistical inference procedure for lattice paths via combinatorial enumerations. The lattice path method is applied to the topological characterization of the protein structures of the COVID-19 virus. We demonstrate that there are topological changes during the conformational change of spike proteins.
\end{abstract}

\section{Introduction}
\vspace{-5pt}
Despite its rigorous mathematical foundation developed for two decades starting with study \cite{edelsbrunner.2000},  persistent homology still suffers from numerous statistical and computational problems. It has not yet become a standard tool in medical imaging. Persistent homology has been applied to a wide variety of data including brain networks \cite{chung.2019.NN}, protein structures \cite{gameiro.2015}, RNA viruses \cite{chan.2013} and molecular structures \cite{meng.2021}. However, most of these methods only serve as exploratory tools that provide descriptive summary statistics rather than formal inference. The main difficulty is due to the heterogeneous nature of topological features, which do not have a one-to-one correspondence across persistent diagrams. Motivated by these challenges, we propose a more principled  topological inference procedure through lattice paths. 

Lattice paths are widely studied algebraic objects in combinatorics and may have potential applications in persistent homology \cite{billera.2001,chung.2019.NN,simion.2000,stanley.1999}. Here, we propose to use the lattice path approach in computing probabilistic statements about the similarity of two persistent diagrams. This is often needed to produce some baseline quantitive measure, such as the $p$-value, commonly used in biomedical research \cite{chazal.2013,chung.2019.NN}. Existing methods for computing $p$-values usually rely on approximate time consuming resampling techniques: jackknife, bootstrap and the permutation test \cite{ahmed.2014,chung.2019.NN}. However, our approach is analytic and  thus compute the \underline{exact} probability without computational burden.

\begin{figure}[t]
\center
  \includegraphics[width=1\textwidth]{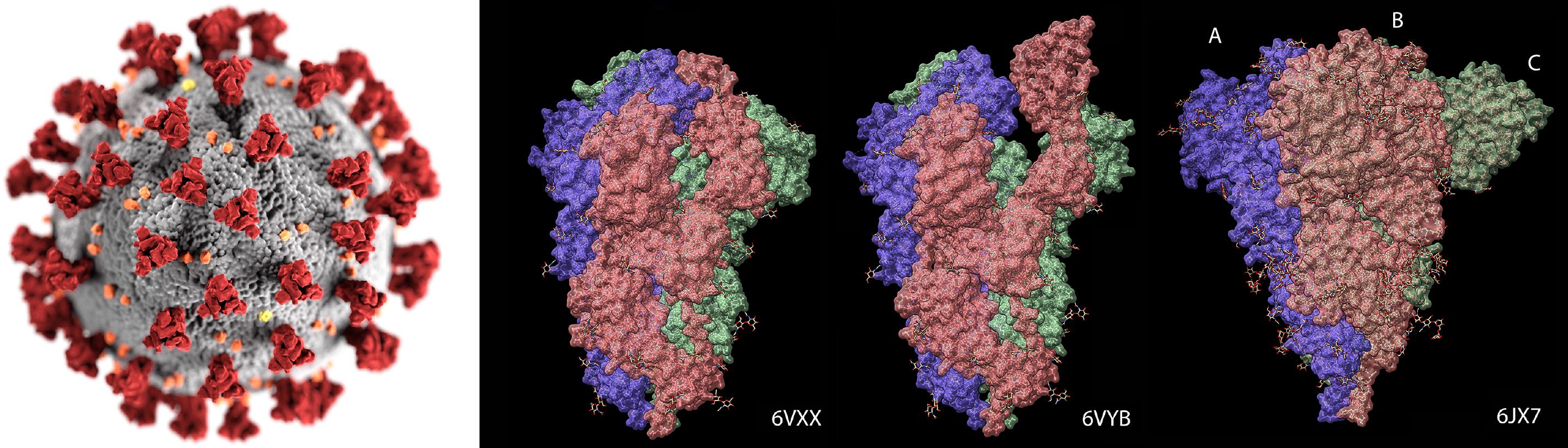}
  \caption{\small Left: COVID-19 virus with spike proteins (red). Right: Spike proteins of the three different corona viruses. The spike proteins consist of three similarly shaped interwinding substructures identified as A (blue), B (red) and C (green) domains.}
  \label{fig:covid}
\end{figure}

The main contributions of this paper are the following: (1) a new data representation via Dyck and lattice paths; (2) the analytic approach for computing probabilities without resampling and significantly reducing run time;  (3) the first topological study on the shape of COVID-19 virus spikes proteins. The proposed lattice path method was used in differentiating the conformational changes of the COVID-19 virus spike proteins that is needed for the virus  to penetrate host cells (Figure \ref{fig:covid}). This demonstration is particularly relevant due to the potential for advancing vaccine development and the current public health concern \cite{walls.2020,cai.2020}.

\section{Methods}
\noindent {\bf Simplicial homology}.
High dimensional objects, such as proteins and molecules, can be modeled as a point cloud data $V$ consisting of $p$ number of points (atoms) indexed as $V=\{1, 2, \cdots, p\}$. Suppose that the distance $\rho_{ij}$ between points $i$ and $j$ satisfies the metric properties. For proteins, we can simply use the Euclidean distance between atoms in a molecule. Then $\mathcal{X} = (V, \rho), \rho=(\rho_{ij})$ is a metric space where we can build a filtration necessary for persistent homology. If we connect points following some criterion on the distance, they will form a simplicial complex which will follow the topological structure of the molecule \cite{edelsbrunner.2010,hart.1999,zomorodian.2009}.  The $k$-simplex  is the convex hull of $k+1$ points in $V$.  A simplicial complex is a finite collection of simplices such as points (0-simplex), lines (1-simplex), triangles (2-simplex) and higher dimensional counterparts. In particular, the {\em Rips complex} $\mathcal{X}_{\epsilon}$ is a simplicial complex, whose $k$-simplices are formed by $(k+1)$ points which are pairwise within distance $\epsilon$ \cite{ghrist.2008}. The Rips complex induces a hierarchical nesting structure called  the Rips  filtration
$\mathcal{X}_{\epsilon_0} \subset \mathcal{X}_{\epsilon_1}\subset \mathcal{X}_{\epsilon_2}\subset \cdots $
for filtration values $0=\epsilon_{0} < \epsilon_{1} < \epsilon_{2} < \cdots$. The filtration is quantified through {\em $k$-cycles} where 0-cycles are the connected components, 1-cycles are loops while 2-cycles are 3-simplices (tetrahedron) without interior.  During the Rips filtration, the $i$-th $k$-cycles are born at filtration value $b_i$ and die at $d_i$. The collection of all the paired filtration values $\{ (b_1, d_1), \cdots, (b_q, d_q) \}$ displayed as scatter points in 2D plane is called the {\em persistent diagram}.\\ 

\noindent {\bf Dyck paths}. The first step in the proposed {\em lattice path method} is to sort the set of all the birth and death values in the filtration as 
order statistics $c: c_{(1)} < c_{(2)} < \cdots < c_{(2q)},$
where $c_{(i)}$ is one of the birth or death values. The subscript $_{(i)}$ denotes the $i$-th smallest value. We will simply call such sequence as the {\em birth-death process}. 
Every possible valid sequence of birth and death values can be viewed as forming a probability space, where each valid sequence is likely to happen with equal probability. 
During the filtration, the sequence of birth and death occurs somewhat randomly but still maintains a specific pairing structure. 

 \begin{figure}[t]
\center
  \includegraphics[width=1\textwidth]{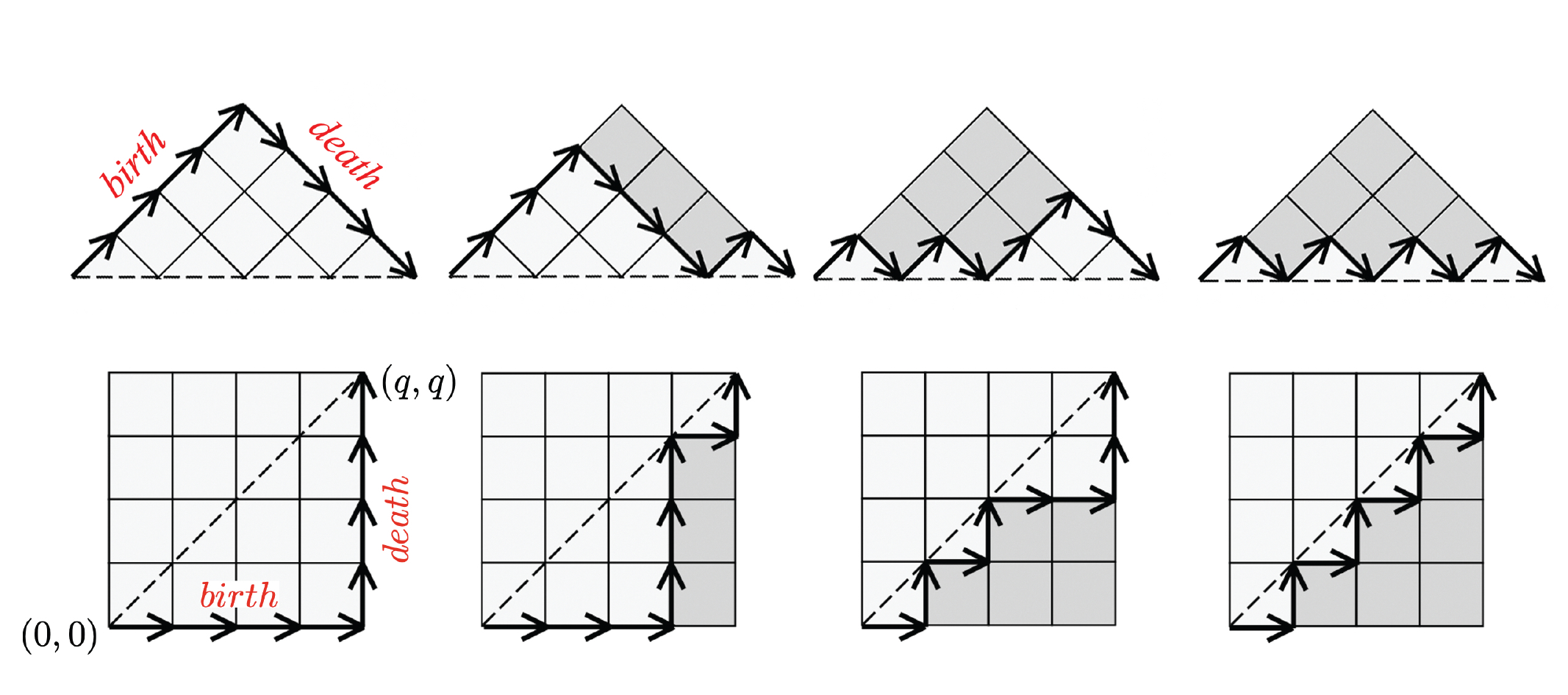}
  \caption{\small Top: 4 different Dyck paths out of 14 possible paths for $q=4$. Bottom: corresponding lattice paths.}
  \label{fig:latticepath}
\end{figure}

There exists a one-to-one relation between the ordering information and Dyck paths if we identify births with $ \nearrow$ and deaths with $\searrow$ \cite{billera.2001,simion.2000}. If we trace the arrows, we obtain the Dyck path (Figure \ref{fig:latticepath}) \cite{stanley.1999}.  A valid Dyck path always starts at $y=0$ and ends at $y=0$. At any moment during the filtration, a Dyck path cannot go below $y=0$. The total number of Dyck paths 
is called the Catalan number $\kappa_p = \frac{1}{q+1} {2q \choose q}$. The first few Catalan numbers 
are $\kappa_1 = 1, \kappa_2 =2, \kappa_3 = 5$ and $\kappa_4 = 14$.  
More rapid changes in the direction of Dyck paths imply more fleeting fluctuations which are indicative of smaller topological signals. 
Less fluctuations indicate larger persistence and thus larger topological structures. The first path in Figure \ref{fig:latticepath} has larger 
persistence while the last path has smaller persistence. \\

\noindent {\bf Lattice paths}. If we rotate the Dyck paths clockwise at 45$^{\circ}$ and flip vertically, we obtain equivalent {\em monotone lattice paths} consisting of a sequence of $\rightarrow$ (uparrow) and $\uparrow$ (downarrow). Figure \ref{fig:latticepath} displays corresponding monotone {\em lattice paths} between $(0,0)$ and $(q,q)$  where the path does not pass above the diagonal line $y=x$ \cite{stanley.1999}. During the filtration, there cannot be more deaths than births and thus the path must lie below the diagonal line. The total area below Dyck paths can be used to quantify the Dyck paths  \cite{chapman.1999}. Since the area below a Dyck path is equivalent to  $q^2/2$ subtracted by the total area of boxes below the corresponding lattice path,
we will simply use lattice paths for quantification. If we tabulate how the area of boxes change over the $x$-coordinate in a lattice path, it is monotone. In the first path in  Figure \ref{fig:latticepath}, the number of boxes below the first and the last lattice paths are $(0,0,0,0)$ and $(0,1,2,3)$. The area below the path is related to persistence. A barcode with smaller persistences (last path in Figure \ref{fig:latticepath}) will have more boxes (dark gray boxes) while longer persistences will have fewer boxes (first path in Figure \ref{fig:latticepath}). Given the sequence of heights of piled-up boxes, we can recover the corresponding lattice path by tracing the outline of boxes. We can further recover the original pairing information about births and deaths. In the Rips filtration for 0-cycles, persistent diagrams line up vertically as $(0, d_{(i)})$. We simply augment them as $ ( (i-1) \delta, d_{(i)})$ for sufficiently small $\delta$.

The lattice and Dyck path representations only encode the ordering information about how births and deaths are paired, and do not encode the actual filtration values. This is remedied by adaptively weighting the length of arrows in lattice paths. We sort the set of birth values $b_i$ and death values $d_i$ as the order statistics:
$$b_{(1)} < b_{(2)} < \cdots < b_{(q-1)} < b_{(q)}, \quad d_{(1)} < d_{(2)} < \cdots < d_{(q-1)} < d_{(q)}.$$
We start at origin $(0,0)$. When we encounter a birth $b_{(i)}$, we take the horizontal step to $b_{(i)}$. When we encounter a death $d_{(i)}$, take the vertical step to $d_{(i)}$ (Figure \ref{fig:latticepath}). 
 The weighted lattice paths contains the same topological information as the original persistent diagram. \\

\begin{figure}[t]
\centering
\includegraphics[width=1\linewidth]{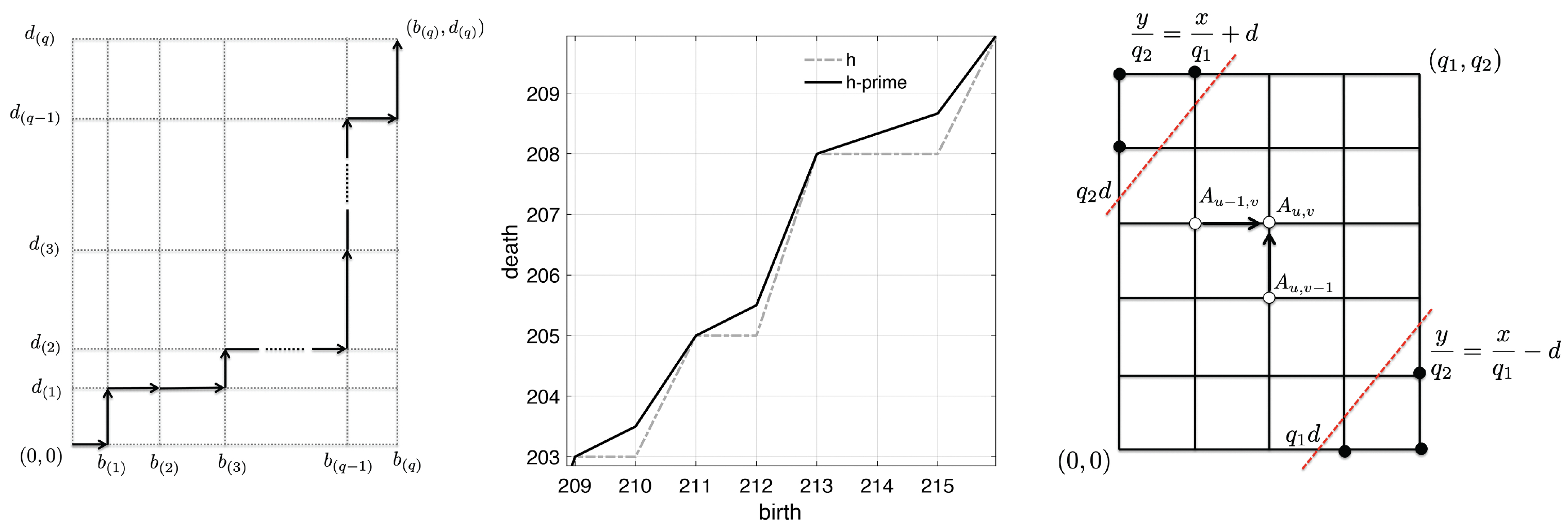}
\caption{\small Left: Weighted lattice path equivalent to a persistent diagram. Middle: The area of boxes below lattice paths $h$ (dotted line) is made into strictly increasing to $h'$ (solid line). Right: The problem of lattice path enumeration between $(0,0)$ and $(q_1,q_2)$ with the constraint $|x/q_1-y/q_2| <d$.}
\label{fig:hprime}
\end{figure}

\noindent {\bf Exact topological inference}. 
 Using the weighted lattice paths, we can provide the probabilistic statement about how close two  birth-death processes are, which can be used for topological inference. For this, we need the transformation $\phi$:
 
\begin{theorem}\label{thm::phi} 
There exists a one-to-one map from a birth-death process to a monotone function $\phi$  with $\phi(0)=0$ and $\phi(1) =q$. 
\end{theorem}
{\em Proof.} We explicitly construct such a function $\phi$. Consider the sequence of areas of boxes  as we traverse the weighed lattice path: $ h: h_1 \leq h_2 \leq  \cdots \leq h_{q},$
where $h_{i+1} = (b_{(i+1)} - b_{(i)})(d_{(i+1)} - d_{(i)})$ is the area of $i$-th box with $h_1 =0$. The areas $h$ may not strictly increase (Figure \ref{fig:hprime}). If births occurs $r$ times sequentially in the birth-death process,  $h$ will have $r$ repeated identical areas $h_i, \cdots, h_i$ as a subsequence. To make the subsequence  strictly increasing, we simply add a sequence of strictly increasing small numbers $\delta (0, 1, 2, \cdots, r-1)$ to the repetition with  $\delta \leq \frac{1}{r}$ (Figure \ref{fig:hprime}). Denote the transformed sequence as $h': h_1' < \cdots < h_q'$. Then $\phi(t)$ is given as a step function
\bq 
\phi(t) = 
\begin{cases}
0 \quad \mbox{ if }  t \in [0,  \frac{h_1'}{q})\\
j  \quad \mbox { if } t \in [\frac{h_{j}'}{q}, \frac{h_{j+1}'}{q})  \; \mbox{ for } j= 1, \cdots, q-1\\
q \quad \mbox{ if } t \in [\frac{h_q'}{q},1] 
\end{cases}. \label{eq:step} 
\eq
From $\phi(t)$, the original sequence $h$ and  the original birth-death process can be recovered exactly. Such map from a birth-death process to $\phi$ is one-to-one. \hfill   $\square$

\begin{figure}[t]
\center
  \includegraphics[width=1\textwidth]{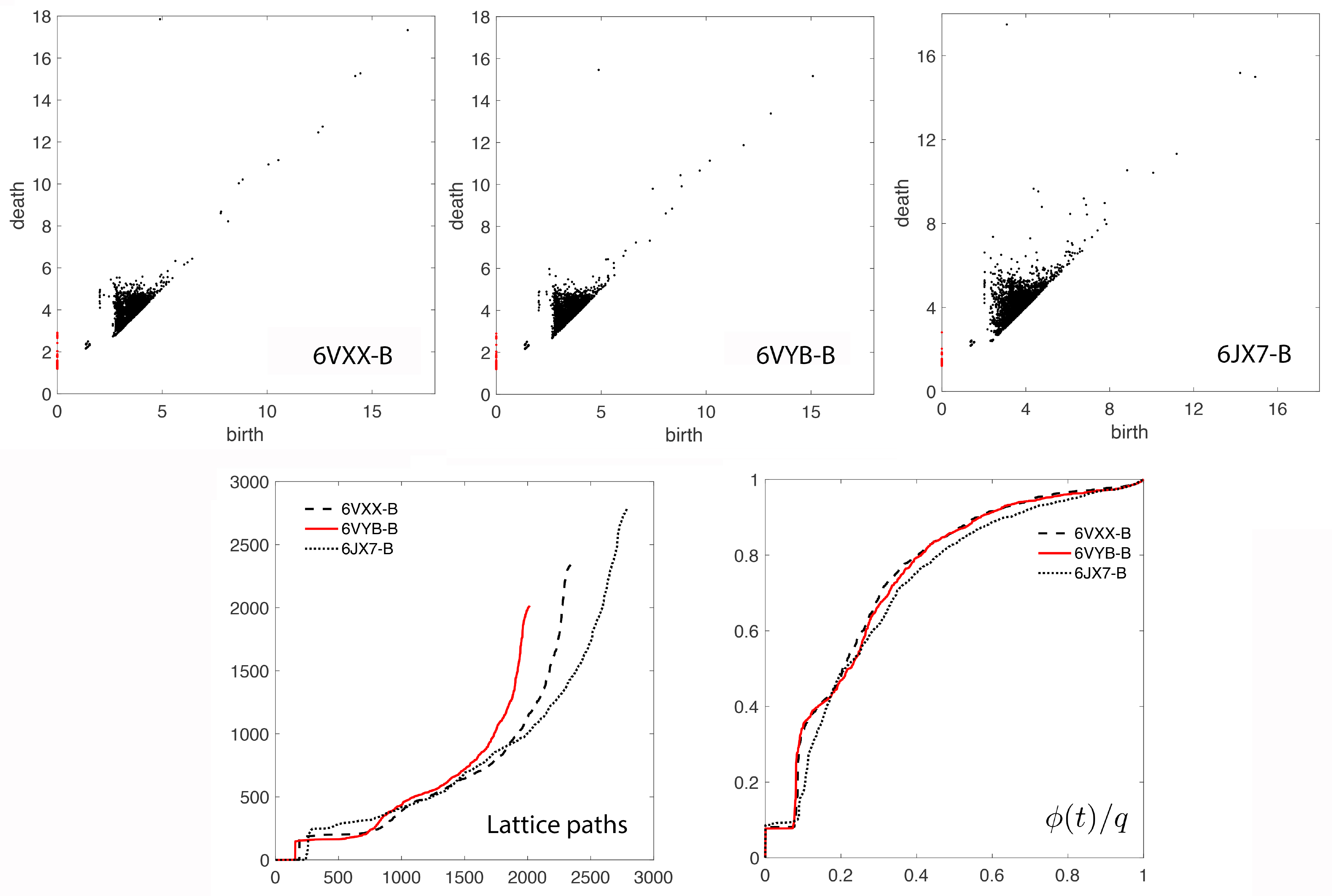}
  \caption{\small Top: persistent diagrams of three different spike proteins. The red dots are 0-cycles and the black dots are 1-cycles. 
  The units are in $\text{\normalfont\AA}$ (angstrom). Bottom: the corresponding lattice paths and normalized step functions $\phi(t)/q$.}
  \label{fig:PD}
\end{figure}

Note $ \| h - h ' \|_2 \to 0$ as $\delta \to 0$. So by making $\delta$ as small as possible, we can construct a strictly monotone $h'$ to be arbitrarily close to $h$. The normalized step function $\phi(t)/q$ can be viewed as an {\em empirical cumulative distribution} and many statistical tools for analyzing distributions can be readily applied. Figure \ref{fig:PD}-bottom displays the lattices paths and the normalized step functions of 1-cycles corresponding to the spike proteins used in the study.

With monotone function $\phi$, we are ready to test the {\em topological equivalence} of two birth-death processes:
\bq C^1 : c^1_1 < c^1_2 < \cdots < c^1_{q_1}, \quad C^2 :  c^2_1 < c^2_2 < \cdots < c^2_{q_2}.
\eq
Let $\phi_1$ and $\phi_2$ be the step functions corresponding to $C^1$ and $C^2$. The topological distance
$$D(\phi_1,\phi_2) = \sup_{t \in [0,1]} \Big| \frac{\phi_1 (t)}{q_1} - \frac{\phi_2 (t)}{q_2} \Big|$$
will be used as the test statistic for testing the equivalence of $C^1$ and $C^2$. The normalizing denominators $q_1$ and $q_2$ ensures that the value of step  
functions are in $[0,1]$. The statistic $D(\phi_1,\phi_2)$ is the upper bound of area difference under $\phi_1(t)/q_1$ and $\phi_2(t)/q_2$:
$$\int_0^1 \Big| \frac{\phi_1 (t)}{q_1} - \frac{\phi_2(t)}{q_2} \Big| \; dt \leq D(\phi_1,\phi_2).$$
\begin{theorem} Under the null hypothesis of the equivalence of $C^1$ and $C^2$,
$$P( D(\phi^1,\phi^2) \geq d )   = 1 - \frac{A_{q_1,q_2}}{{q_1 + q_2 \choose q_1
}},$$
where $A_{u,v}$ satisfies $A_{u,v} = A_{u-1,v} + A_{u, v-1}$
with the boundary condition $A_{q_1,0}=A_{0,q_2}=1$ within the band $|u/q_1 - v/q_2| < d$. 
\label{thm:null}
\end{theorem}

\noindent {\em Proof.} The statement can be proved similarly as the combinatorial construction of the Kolmogorov-Smirnov test \cite{bohm.2010,gibbons.1992,chung.2019.NN}. First, we combine monotonically increasing sequences $C^1$ and $C^2$ and sort them into a bigger monotone sequence of size $q_1 + q_2$. Then, we represent the combined sequence as the sequence of $\rightarrow$ and $\uparrow$ respectively depending on if they are coming from $C^1$ or $C^2$. Under the null, there is no preference and they equally likely come from $C^1$ or $C^2$. If we follow the sequence of arrows, it forms a monotone lattice path from $(0,0)$ to $(q_1,q_2)$.  In total, there are  ${q_1 + q_2 \choose q_1}$ possible equally likely lattice paths that forms the sample space. 
From Theorem \ref{thm::phi},  the values of $\phi_1(t)$ and $\phi_2(t)$ are integers from 0 to $q$. Then it follows that  
$$P (D \geq d ) = 1 - P ( D_q < d ) = 1 - \frac{A_{q_1,q_2}}{{q_1 + q_2 \choose q_1}},$$
where $A_{u,v}$ is the total number of valid paths from $(0,0)$ to $(u,v)$ within dotted red lines in Figure \ref{fig:hprime}.
$A_{u,v}$ is iteratively computed using $A_{u,v} = A_{u-1,v} + A_{u, v-1}.$
with the boundary condition $A_{u,0}=A_{0,v}=1$ for all $u$ and $v$. \hfill $\square$

Computing $A_{q_1,q_2}$ iteratively requires at most $q_1 \cdot q_2$ operations while the permutation test will cause a  computational bottleneck for large $q_1$ and $q_2$. Thus, the proposed {\em lattice path method} computes the exact $p$-value substantially faster than the permutation test. 
Since most protein molecules consist of thousands of atoms, $q_1$ and $q_2$ should be sufficiently large to apply the asymptotic  \cite{gibbons.1992,smirnov.1939,chung.2017.IPMI}: 

\begin{theorem} 
\label{theorem:lim2}
$\lim_{q_1,q_2 \to \infty}  P\Big( \sqrt{\frac{q_1q_2}{q_1 + q_2}} D \geq  d  \Big)  = 2 \sum_{j=1}^{\infty} (-1)^{j-1}e^{-2j^2d^2}.$
\end{theorem}

Subsequently, the $p$-value under null is given by
$$\mbox{$p$-value} = 2 e^{-d_{o}^2} - 2e^{-8d_{o}^2} + 2 e^{-18d_{o}^2} \cdots,$$ where $d_{o}$ is the observed value of 
$\sqrt{\frac{q_1q_2}{q_1 + q_2}}D$.  
Computing the $p$-value through Theorem \ref{theorem:lim2} mainly requires sorting, which has the runtime of  $\mathcal{O} \big( q \log q \big)$ for $q=q_1=q_2$. On the other hand, the traditional permutation test requires computing the distance for ${2q \choose q}$ possible permutations, which is asymptotically  $\mathcal{O}(4^q /\sqrt{\pi q} )$ \cite{chung.2019.CNI,feller.2008}. For thousands of atoms, the total number of permutations is  too large to compute. Thus, only a small fraction of  randomly generated permutations are used in the traditional permutation test  \cite{chung.2017.IPMI,chung.2018.MICCAI,chung.2019.CNI,nichols.2002,thompson.2001,zalesky.2010}. Even if we use hundreds of thousands permutations, the traditional permutation test still takes a significant computational effort.
Further, as an approximation procedure, the standard  permutation test does not perform better than the exact topological inference, which gives the mathematical ground truth. This is demonstrated in Table 1 in the simulation study \cite{chung.2018.MICCAI}.

\section{Application: spike proteins of COVID-19 virus}
\vspace{-5pt}
The proposed lattice path  method is used to study the topological structure of the severe acute respiratory syndrome coronavirus 2 (SARS-Cov-2), which is often called COVID-19. Since the start of the global pandemic (approximately December 2019), COVID-19 has already caused 3.85 million deaths in the world as of June 2021. The COVID-19 virus is specific member of a much broader coronavirus family, which all have spike proteins surrounding the spherically shaped virus similar to the sun's corona. The glycoprotein spikes bind with receptors on the host's cells and consequently cause  severe infection. The atomic structure of spike proteins can be determined through the cryogenic electron microscopy (cryo-EM) \cite{walls.2020,cai.2020}. Figure \ref{fig:covid}-left illustrates spike proteins (colored red) that surround the spherically shaped virus. Each spike consists of three similarly shaped protein molecules with rotational symmetry often identified as A, B and C domains.  The spike proteins have two distinct conformations identified as {\em open} and {\em closed} states, where the domain's opening is necessary for interfacing with the host cell's surface for membrane fusion and viral entry (Figure \ref{fig:covid}-right). Indeed, most current vaccine efforts focus on preventing the open state from interfacing with the host cell. Hence, this line of research is of prime importance in vaccine development and therapeutics \cite{cai.2020}.

In this study, we analyzed the spikes of three different coronaviruses identified as 6VXX, 6VYB \cite{walls.2020} and 6JX7 \cite{yang.2020}. The 6VXX and 6VYB are respectively the closed and open states of SARS-Cov-2 from human while 6JX7 is feline coronavirus (Figure \ref{fig:covid}).  All the domains of 6VXX have exactly 7604 atoms and are expected to be topologically identical. Applying the lattice path method to 1-cycles, we tested the topological equivalence of the B-domain and the A- and C-domains within 6VXX. The normalized step functions are  almost identical and the observed topological distances are 0.0090 and 0.0936, which give the $p$-value of  1.00 each. As expected, the method concludes that they are topologically equivalent. Figure \ref{fig:PD}-bottom displays  the lattice paths and the normalized step functions for domain B of 6VXX. The plots for other domains are visually almost inseparable and hence not shown. 
The closed domain B of 6VXX  is also compared against the open domain B  of 6VYB.  The open  state has a significantly reduced number of atoms at 6865 due to the conformational change that may change the topology as well. The observed topological distance is 0.20 and with an extremely small  $p$-value of $8.1123 \times 10^{-38}$ which strongly suggests evidence for topological change. The persistent diagrams of both closed and open states are almost identical in smaller birth and death values below 6 $\text{\normalfont\AA}$ (angstrom)  (Figure \ref{fig:PD}-top). The major difference is in the scatter points with larger brith and death values. The  lattice path method confirms that the local topological structures are almost identical while the global topological structures are different.

The domain B of 6VXX is also compared against the domain B of  feline coronavirus 6JX7 consisting of 9768 atoms. Since 6JX7 is not from human, it is expected that they are different. The topological distance is 0.9194 and $p$-values is $0.00 \times 10^{-38}$ confirming that the topological nature of spikes are different. This shows the biggest  difference among all the comparisons done in this study. 
 The Matlab codes and data used for the study are available at \url{http://www.stat.wisc.edu/~mchung/TDA}.

\section{Conclusions}\vspace{-5pt}
In this paper, we proposed a new representation of persistent diagrams using lattice paths. The novel representation enable us to perform the statistical inference combinatorially by enumerating every possible valid lattice paths analytically. The  proposed lattice path method is subsequently used to analyze the coronavirus spike proteins. The normalized step functions $\phi(t)/q$ for all the spike proteins show fairly stable consistent global monotone pattern but with localized differences. We demonstrated the  lattice path method has the ability to statistically discriminate between the conformational changes of the spike protein that are needed in the transmission of the virus. We hope that the our new representation enables scientists in their effort to automatically identify the different types and states of coronaviruses in a more principled manner. 

\section*{Acknowledgement}
The illustration of COVID-19 virus (Figure \ref{fig:covid} left) is provided by Alissa Eckert and Dan Higgins of Disease Control and Prevention (CDC), US. The proteins 6VXX and 6VYB are provided by Alexander Walls of University of Washington. The protein 6JX7 is provided by Tzu-Jing Yang of National Taiwan University. Figure 2-left is modified from an image in Wikipedia. This study is supported by NIH EB022856 and EB028753, NSF MDS-2010778, and CRG from KAUST. 

\bibliographystyle{plain}
\bibliography{reference.2021.07.27}

\end{document}